\documentclass[runningheads]{llncs}

\usepackage[T1]{fontenc}
\usepackage{lmodern}

\usepackage[pdftex]{graphicx}
\usepackage[caption=false,font=normalsize,labelfont=sf,textfont=sf]{subfig}
\usepackage[ruled,vlined]{algorithm2e}
\usepackage{url}
\usepackage{algorithmic}

\usepackage[utf8]{inputenc} % allow utf-8 input
\usepackage[T1]{fontenc}    % use 8-bit T1 fonts
\usepackage{hyperref}

\usepackage{nicefrac} 
\usepackage{microtype} 
\usepackage{xcolor}
\usepackage{bm}
\usepackage{ulem}
\usepackage{comment}
\newcommand\redsout{\bgroup\markoverwith{\textcolor{red}{\rule[0.5ex]{2pt}{0.4pt}}}\ULon}

\begin{document}
\title{{ExMo}: {\textbf{\textit{Ex}}}plainable AI {\textbf{\textit{Mo}}}del using Inverse Frequency Decision Rules}
%Inverse Frequency Decision Rules for Interpretable Machine Learning
%
%\titlerunning{Abbreviated paper title}
% If the paper title is too long for the running head, you can set
% an abbreviated paper title here
%
\author{Pradip Mainali\inst{1} \and
Ismini Psychoula\inst{1} \and
Fabien A. P. Petitcolas\inst{2}}
\authorrunning{P. Mainali et al.}
% First names are abbreviated in the running head.
% If there are more than two authors, 'et al.' is used.
%
\institute{OneSpan, Cambridge, The United Kingdom \and
OneSpan, Brussels, Belgium \\
\email{{first}.{last}@onespan.com}
}
\maketitle              % typeset the header of the contribution
\begin{abstract}
In this paper, we present a novel method to compute decision rules to build a more accurate interpretable machine learning model, denoted as ExMo. The ExMo interpretable machine learning model consists of a list of IF...THEN... statements with a decision rule in the condition. This way, ExMo naturally provides an explanation for a prediction using the decision rule that was triggered. ExMo uses a new approach to extract decision rules from the training data using term frequency-inverse document frequency (TF-IDF) features. With TF-IDF, decision rules with feature values that are more relevant to each class are extracted. Hence, the decision rules obtained by ExMo can distinguish the positive and negative classes better than the decision rules used in the existing Bayesian Rule List (BRL) algorithm, obtained using the frequent pattern mining approach. The paper also shows that ExMo learns a qualitatively better model than BRL. Furthermore, ExMo demonstrates that the textual explanation can be provided in a human-friendly way so that the explanation can be easily understood by non-expert users. We validate ExMo on several datasets with different sizes to evaluate its efficacy. Experimental validation on a real-world fraud detection application shows that ExMo is $\approx$20\% more accurate than BRL and that it achieves accuracy similar to those of deep learning models.
\keywords{Interpretable  \and Explainable \and Decision Rules.}
\end{abstract}
\section{Introduction}
Artificial intelligence (AI) and machine learning (ML) show promising results in many application domains. Accenture projects that AI will boost the global economy by 35\% by 2035~\cite{accenture_2035,accenture_xai} and that individuals will be 40\% more productive by employing AI~\cite{accenture_2035}. Currently, the adoption of AI is mainly for low-stake applications such as recommendation of products in e-commerce. However, in high-stake applications such as finance, medicine, justice, etc., the adoption of AI is not taking place at the same speed. One of the hurdles is the black-box nature of machine learning and deep learning methodologies. Complex deep learning models can achieve higher accuracy, but they are difficult to interpret and understand the rationale behind their decision making. In high-stake applications, understanding the rationale behind decision making is important to build trust in the system. For example, doctors may want to scrutinize the decision of AI, if AI predicts surgery. Also, there is a legal obligation mentioned in the EU General Data Protection Regulation (GDPR) such as `right to explain'~\cite{gdpr_right_to_explain}, hence autonomous systems are required to provide explanations to meet the legal compliance. 

The research directions to deal with this problem are Explainable AI and Interpretable AI. In the Explainable AI approach, the predictions are made using a complex black-box model, and a second simpler model is used to help explain what the black-box model is doing locally~\cite{molnar2019,peeking_xai}. Well known model-agnostic approaches are SHAP~\cite{shap} and LIME~\cite{lime}, which provide explanations for any machine learning model. Another method designed specifically for deep neural networks is called Integrated Gradients (IG)~\cite{ig}. Interpretable AI models can be directly inspected and interpreted by human experts~\cite{molnar2019}. A study carried out in~\cite{stop_explaining} strongly recommends using interpretable models for high-stake applications. It used to be a common myth that interpretable models were less accurate, however, recent advances with rule list algorithms have led to the ability to build more accurate interpretable models ~\cite{stop_explaining}. Our paper aims to improve the accuracy of interpretable models even further. 

This paper proposes a new approach to compute decision rules and an interpretable model is built using them. The method is named ExMo. We compare ExMo with the existing BRL algorithm~\cite{brl}, which uses frequent pattern mining (FPGrowth)~\cite{fpgrowth}, and is commonly used for interpretability, and show that the decision rules provided by ExMo offer greater accuracy and quality. Basically, ExMo advances the research work in BRL further by adopting a new approach for computing the decision rules. The ExMo model is composed of a list of IF...THEN... statements with a decision rule in the `IF' condition and the `THEN' part providing the prediction score. Consequently, the model naturally provides an explanation because the features in the decision rule are the main reasons for that prediction. Moreover, the decision rule resembles very highly the human decision-making process where the human reasoning mostly starts with `IF' and conditions, hence the model is highly interpretable~\cite{molnar2019}. To extract the decision rules from the training data, we use features extracted using the term frequency-inverse document frequency (TF-IDF) technique from natural language processing. First, two document classes (i.e., fraud and non-fraud) are created by converting the fraud and non-fraud samples into a text representation, and then the text represented samples are assembled to create two documents (one for each class). TF-IDF features are based on the number of times that features have occurred in each class, which is the term frequency (TF). After that, the inverse document frequency (IDF) is computed: it is a weight computed for the same features in TF and whether these features occur in both the documents or not. If the features occur in both documents, then the weights are small and vice-versa. Finally, TF-IDF features are computed by multiplying both the terms. In essence, IDF de-emphasizes the decision rules with the features that occur in both document classes. Consequently, our approach provides sharper decision rules than the decision rules extracted using frequent pattern mining such as FPGrowth~\cite{fpgrowth} as used in the original BRL algorithm~\cite{brl}. In the frequent pattern mining approach, decision rules are computed independently for each class and hence the decision rules are blurry and make the interpretable model less accurate.

The main advantages of the ExMo interpretable model are that the explanation is obtained at no additional computational cost, the ExMo model can be audited by the regulators, and the explanation in the textual form is also provided for non-expert users. Model-agnostic methods such as SHAP~\cite{shap} and LIME~\cite{lime} require perturbation of the training data and need to make predictions for all perturbed data samples to compute the explanation. Consequently, generating the explanation requires expensive computations. The second benefit of ExMo is that the model is readable by human experts, as for example, regulators who can simply read the model in text form to verify for bias and unfairness. AI explainers, such as SHAP, could also be used to provide explanation and auditability to regulators for black-box models. However recently it was shown that AI explainers can be tricked to hide bias~\cite{fooling_lime_shap}, hence AI explainers may not be safe enough to be used in practice. Another advantage of ExMo is that it also provides an explanation in the textual form, that is, in a more human-friendly way suitable for non-expert users. For this, the decision rule is processed further to format the explanation in textual form. This will bring advantages such as ExMo will enable customer support teams to communicate a prediction to the end-user as to why the loan application was declined and help the doctors to scrutinize the decision in greater detail etc.    

The main contribution of this paper is the proposal of a new scheme that uses TF-IDF to extract the decision rules from the training data. We validate the approach with experiments and show that the ExMo interpretable model built using these new decision rules achieves a higher classification accuracy. We also compared ExMo with a deep learning model for fraud detection application, ExMo achieves an accuracy that is close to the deep learning model, while proving an explanation without additional computation time. Furthermore, we also demonstrate that ExMo achieves a qualitatively better model than BRL, thanks to TF-IDF based decision rules. Finally, this paper also demonstrates that ExMo can provide a textual explanation for its prediction, so that non-expert users can also understand the explanation.      

\section{Related Work}
In this section, the machine learning models and explanation approaches are reviewed. We also provide a literature study on how the explanation techniques are being adopted for the development and improvement of machine learning systems. 

{\textbf{Explaining fraud}}: There has been a sharp rise in the literature proposing the use of machine learning in real-life applications. Hence, the integration of explanation techniques is also growing~\cite{ismini_xai_fraud}. Several machine learning models for credit card fraud detection are discussed in~\cite{fraud_imbalance}. A deep learning approach using convolutional neural networks and Long Short-Term Memory Network (LSTM) sequence model was used for fraud detection in~\cite{credit_fraud_detection_2020}. Authors of~\cite{instance_fraud_xai} developed two novel dashboards to visualize features and a Sankey diagram to visualize the decision rules used to explain the prediction made by machine learning models. The authors found that the adoption of this explanation can dramatically speed up the process of filtering potential fraud cases. The SHAP explanation technique was used in~\cite{xai_improve_fraud} to improve a fraud detection model. In their approach, specialists analyzed the output of a fraud detection model with the SHAP technique in each iteration to check whether the model had learned the desired pattern or not. xFraud~\cite{xfraud} is proposed to detect fraud from the graph representation of data for online retail platforms. The explanation is generated using GNNExplainer and the results are fed to the business unit for further processing to understand predictions. A deep learning model is used for fault detection in~\cite{xai_fault_detection_rotary} and SHAP explanation technique was used to understand the prediction of the model.  Finally, in~\cite{watson2021attackagnostic} the authors show how explanation techniques could be used to better identify adversarial attacks and help prevent fraud in the medical domain.

{\textbf{Explainable AI}}: The growing popularity of deep learning methodology and its black-box nature have led to the development of "Explainable AI". Model agnostic and model-specific explanation techniques are two different mechanisms for this. The model agnostic approach creates a second model to explain what the actual deep learning model is doing locally and can explain any machine learning or deep learning model. LIME~\cite{lime} and SHAP~\cite{shap} are popular techniques based on this category. LIME and SHAP provide an explanation based on feature importance, showing the different features that are contributing positively and negatively to the prediction. LIME approximates a local region with an interpretable model such as a linear model. SHAP uses a game-theoretic approach, namely Shapley values to assign credit to input features to generate feature importances. Anchors~\cite{anchors} improved upon LIME by replacing a linear model with a logical rule similar to the decision rule, providing a more selective explanation. Due to data perturbation, the explanation provided by LIME can have slightly different features on each run. To provide a consistent explanation for each run, DLIME~\cite{DLIME} was proposed, which perturbs data based on clusters for more deterministic data perturbation rather than perturbing the whole dataset. GraphLIME~\cite{GraphLIME} is another adaptation of LIME to provide an explanation for models applied to graph-represented data. RESP~\cite{resp} provides an explanation based on causality and has also been validated on credit and fraud datasets. Methods to explain the output of deep neural networks (DNN) have also been developed. DeepLIFT~\cite{Deeplift} computes the importance scores for input features by comparing them with a reference entity. Layer-wise relevance propagation (LRP)~\cite{lrp} explains the classifier's decision by decomposition and by redistributing the prediction backwards through the layers until it assigns a relevance score to each input variable. Integrated Gradients~\cite{ig} accumulate the gradients obtained after perturbing input from the reference value to the input value and provide an explanation as feature importance scores. 

{\textbf{Interpretable AI}}: Interpretable models provide both a prediction and the reason for the prediction. Interpretable models are those that directly build a human-readable model from the data. Recently, interpretable models started to be built using decision rule lists~\cite{scalable_brl,brl,falling_rule_list}. The decision rules are extracted using a frequent pattern mining algorithm. Then the model is built using the decision rules. The decision rules are highly interpretable and naturally provide an explanation for the prediction. The method proposed in~\cite{Okajima_Sadamasa_2019} replaces the last layer of the neural network and predicts over the decision rules, providing both prediction and explanation together. For high-stake applications, the paper~\cite{stop_explaining} suggested following the path of building the interpretable model because it provides stable and truthful explanations. 

\section{Proposed Method}                       
In this section, we explain ExMo in detail on how decision rules are computed from the training data and the interpretable model is built from these decision rules. We also briefly review the Bayesian rule list (BRL) algorithm in Sec.~\ref{subsec:brl}. Following that, our new approach for learning the decision rules using TF-IDF features is explained in Sec.~\ref{subsec:tf_idf_decision_rules}.

\subsection{Review of the Bayesian Rule List (BRL) Algorithm}\label{subsec:brl}
The goal of the BRL algorithm~\cite{brl} is to learn a model which consists of an ordered collection of decision rules $(d=\{{a_i}\}_{i=1}^m)$ as shown in Table~\ref{tab:interpretable_model}, where $m$ is the number of rules in the decision list. The model is specified by a list of IF...THEN... statements with a decision rule ($a_i$) in the condition of the `IF' part and a probability of classification in the `THEN' part as shown in Table~\ref{tab:interpretable_model}.
\begin{table}[h]
    \centering
    \caption{Interpretable model}
    \label{tab:interpretable_model}
    \begin{tabular}{lcll} 
     \hline
     if $a_1$ & then & $y \sim \textrm{Binomial}(\bm{\theta}_1)$, & $\bm{\theta}_1 \sim \textrm{Beta}(\bm{\alpha} + \bm{N}_1)$  \\ 
     else if $a_2$ & then & $y \sim \textrm{Binomial}(\bm{\theta}_2)$, & $\bm{\theta}_2 \sim \textrm{Beta}(\bm{\alpha} + \bm{N}_2)$ \\ 
     .\\
     .\\
     else if $a_m$ & then & $y \sim \textrm{Binomial}(\bm{\theta}_m)$, & $\bm{\theta}_m \sim \textrm{Beta}(\bm{\alpha} + \bm{N}_m)$ \\
     else & & $y$ $\sim \textrm{Binomial}(\bm{\theta}_0)$, & $\bm{\theta}_0 \sim \textrm{Beta}(\bm{\alpha} + \bm{N}_0)$ \\
     \hline
    \end{tabular}
\end{table}
The decision rule $a_i$ is a condition on data sample $x$ that evaluates to true or false. $y$ is a prediction score which is a binomial distribution over labels $\bm{\theta}_i$, which is computed from prior $\bm{\alpha}$ and the likelihood specified with the beta distribution. The vector $\bm{\alpha}=[\alpha_1, \alpha_0]$ has a prior parameter for each class. The notation ${\bm{N}}_i$ is a two-dimensional vector of counts for positive and negative classes. The counts are computed from training data that satisfy the decision rule $a_i$ and none of the previous decision rules from $a_1$ to $a_{i-1}$ in the list. 

In the BRL algorithm, the decision rules are mined from the training dataset using a frequent pattern mining algorithm such as FPGrowth, Apriori, etc.  A frequent pattern mining algorithm computes frequent co-occurrence of feature values. Support is used to measure the occurrence of feature values as below:
\begin{equation}
    \label{eq:fpgrowth_fppattern}
    {\mbox{Support}}(x, f)=\frac{1}{n}\sum_{i=1}^{n} I(x_{i})
\end{equation}
where $f$ is the feature value, $n$ is the number of samples in the training dataset and $I$ the indicator function that returns $1$ if the sample has feature value $f$, otherwise $0$. 

To extract the decision rules for binary classification, the training samples are grouped into positive and negative groups. Then, the frequent pattern mining algorithm is applied on each group to extract decision rules. Later, the decision rules with support values (as in Eq.~\ref{eq:fpgrowth_fppattern}) that are higher than some threshold values are extracted. Once the decision rules from each group are extracted, the decision rules are combined to create a pool of decision rules ($A$).    

The BRL algorithm learns the interpretable model as stated above from the pool of decision rules that was obtained earlier. Bayesian statistics is used to compute the posterior over the decision list, which is computed from the likelihood and prior as below:
\begin{equation}
    \label{eq:posterior_brl}
    p(d | x, y, A, \alpha, \lambda, \eta) \propto p(y|x, d, \alpha) p(d|A, \lambda, \eta)
\end{equation}
where $d$ is an ordered decision list, $x$ the data sample, $y$ the label, $A$ the pool of pre-mined decision rules, $\lambda$ the prior expected length of the decision lists, $\eta$ the prior expected number of conditions in a rule, and $\alpha$ the prior pseudo-count for the positive and negative classes which is fixed at $[1, 1]$ as in~\cite{brl}.

The likelihood for the model is given by:
\begin{equation}
    \label{eq:likelihood_brl}
    p(y|x, d, \alpha) \propto  \prod_{j=0}^{m} \frac{\Gamma{(N_{j, 0} + \alpha_0)} \Gamma{(N_{j,1}+\alpha_1)}}{\Gamma{(N_{j,0}+N_{j,1}+\alpha_0+\alpha_1)}}
\end{equation}
$N_{j,0}$ and $N_{j,1}$ are counts of training observations with labels 0 and 1 respectively that satisfy the decision rule $a_j$ and not $a_1$ to $a_{j-1}$. The likelihood of rule $a_j$ is large if $N_{j,0}$ is large and $N_{j,1}$ is very small or vice versa.  

The prior is given by:
\begin{equation}
    \label{eq:prior_brl}
    p(d|A, \lambda, \eta) = p(m|A, \lambda)\prod_{j=0}^{m}p(c_j|c_{<j}, A,\eta)p(a_j|a_{<j}, c_j, A)
\end{equation}
Here, $c_j$ is the number of features in the rule or cardinality of each decision rule. The first and second terms in Eq.~\ref{eq:prior_brl} are respectively the priors for the number of rules $m$ in the model and the number of features $c_j$ in a decision rule. 

The BRL algorithm can be summarised in the following steps:
\begin{enumerate}
    \item The decision rules ($A$) in Eq.~\ref{eq:posterior_brl} are pre-mined by using the FPGrowth algorithm.
    \item The Decision list $d$ is sampled randomly from the prior distribution.
    \item Markov Chain Monte Carlo (MCMC) sampling with Metropolis-Hastings is used to compute posterior in Eq.~\ref{eq:posterior_brl}, generating a chain of the posterior samples of decision list $d=\{{a_i}\}_{i=1}^m$. The posterior samples in the chain are added by generating the proposal sample $d^*$ by modifying the current $d$ and if the acceptance criteria are satisfied as specified in the Metropolis-Hastings algorithm. The decision list $d$ is modified by adding, moving or removing the decision rule ($a_i$) from the list. The option to move, add or remove the decision rule ($a_i$) in the decision list is chosen randomly. The algorithm is executed for $K$ (e.g., 30,000) iterations. The algorithm runs three different MCMC chains, each initialized randomly.    
    \item Select the decision list from the sampled lists (i.e., chain) with the highest probability according to the posterior distribution.
\end{enumerate}

\subsection{Decision Rules from TF-IDF}\label{subsec:tf_idf_decision_rules}
The quality of the decision rules plays an important role for the BRL algorithm to achieve higher accuracy. Therefore, we borrowed an idea from text mining to extract the best decision rules by using TF-IDF $n$-gram features. To enable the text processing of data, the training data is converted into two documents (i.e., positive and negative classes of documents) by combining all samples from the same class. For example, the text document for a positive class is created by concatenating all samples having positive labels. Before concatenating the samples, the numerical features are also discretized as mentioned in the BRL algorithm. The discretization of samples is done using the algorithm proposed in~\cite{brl_discritization_FayyadI93}. This allows treating the numerical feature as a categorical feature for text processing. The label is also used to find cut points for discretization. Once the data is represented in two text documents, $n$-gram TF-IDF features are computed. `TF' is a term-frequency and measures how important a feature is to a document. `IDF' is an inverse-document-frequency and identifies how important the features are to a specific document. The `IDF' values are smaller for the feature values that are common to both documents. The `IDF' values are larger for feature values that occur only in one document. The final weighting of feature values in a decision rule using the TF-IDF is computed by multiplying both the terms as follows:
\begin{equation}
    \mbox{tf-idf}(f, l) = \mbox{tf}(f, l) \times \mbox{idf}(f)
\end{equation}
where
\begin{equation}
    \mbox{idf}(f) = 1 + \log \frac{1+N}{1+\mbox{df}(f)},
\end{equation}
$l$ is the document class, $N$ is the number of documents ($N=2$ in our case), and where $\mbox{df}(f)$ is the number of documents that contain a term or feature value $f$. We can observe that `TF' has a similar meaning to that of `Support' in Eq.~\ref{eq:fpgrowth_fppattern}. In our approach, the term `IDF' helps to emphasize the decision rules with feature values that occur in one document class only and de-emphasize the decision rules with feature values that occur in both the classes. Hence, the decision rules computed by our method are better in quality than those obtained by the frequent pattern mining algorithms as used in the original BRL algorithm. 

\begin{algorithm}[!h]
\caption{Algorithm for computing decision rules}
\label{alg:decision_rules}
\begin{algorithmic}[1]
\STATE Initialize cardinality of decision rules, i.e. $n$-gram range setting for TF-IDF
\STATE Initialize maximum number of permutations $P$
\STATE Pool of decision rules $A=$~[ ]
\FOR{$t=1...P$ } 
\STATE Permute position of feature columns in tabular data
\STATE Convert samples to text and create positive and negative document classes
\STATE Compute $n$-gram TF-IDF features 
\STATE Get top-$k$ n-gram features from each document class
\STATE Append the top-$2k$ features to $A$
\ENDFOR 
\STATE Randomly shuffle the decision rules $A$
\end{algorithmic}
\end{algorithm}

The algorithm to compute the decision rules using TF-IDF $n$-gram features is given in Algorithm~\ref{alg:decision_rules}. In the first step, the algorithm is initialized for cardinality and the number of permutations ($P$) to be performed. The cardinality mentioned in Eq.~\ref{eq:prior_brl} corresponds to the maximum number of features in a decision rule. This also corresponds to a maximum value in $n$-gram range settings for the TF-IDF algorithm. For example with the $n$-gram range setting $(3, 5)$, a decision rule with a maximum number of 5 features are computed. The next parameter is $P$. We permute the position of the feature columns in the tabular data to compute the decision rules with different n-gram features. The number of times that the position of feature columns are permuted is defined by $P$. In each iteration from steps 3 to 9, the position of feature columns in tabular data is permuted and the data is converted to two document classes as discussed earlier. The $k$ top $n$-gram features from each class are extracted. In total, $2\times k$ decision rules are extracted from each iteration. The process is repeated for $P$ number of permutations. In the end, the algorithm outputs $2k \times P$ decision rules. The pool of decision rules is randomly shuffled. This pool of decision rules is given as input to the BRL algorithm to learn the interpretable model. 

\section{Experimental Results}\label{sec:experiment}
In this section, we provide the experimental results of classification accuracy, comparison of the models, comparison of explanations generated by different algorithms, and execution time to compute explanations. We also provide details of the datasets and the features present in the dataset. We compare the explanations generated by our algorithm with BRL, LIME~\cite{lime}, SHAP~\cite{shap}, Anchor~\cite{anchors} and IG~\cite{ig}. 

\subsection{Datasets}\label{sec:exp_datasets}
We evaluate the algorithms on small and large datasets to evaluate the efficacy of ExMo. The small dataset consists of a few samples and features. The large dataset consists of a huge number of samples and features so that the scalability of ExMo can be evaluated. The details of the datasets are given below. We also provide some information regarding the features in the dataset as well so that the explanations given by the algorithms can be illustrated better.    
\subsubsection{Small Dataset:} We used Diabetes~\cite{diabetes_dataset}, Default Credit Card Payment~\cite{defaultcreditcard_dataset} and Income over 50K (or adult)~\cite{adult50k} datasets, having 769, 30,000 and 48,842 samples, respectively. In terms of the number of features, the datasets contain 8, 23 and 14, respectively. We used 80\% of data for training and 20\% of data for testing. We also use the diabetes dataset to compare the models and to compare textual explanations, hence we provide more information regarding the features. The features in the diabetes dataset are: the number of times pregnancy occurred (times{\_}pregnant), plasma glucose level after 2 hours (plasma{\_}glucose), diastolic{\_}blood{\_}pressure, tricep{\_}skin{\_}fold{\_}thickness, serum{\_}insulin, body{\_}mass{\_} index, diabetes{\_}pedigree{\_}function and age. 

\subsubsection{Large Dataset:}
We carried out the experiment on the large dataset so that the experiment demonstrates that ExMo can be applied in the large dataset as well. For this, we selected the IEEE-CIS Fraud Detection dataset~\cite{ieee_fraud}, which is a real-world dataset consisting of 500K samples and 433 features for online payments collected by Vestas Corporation. The dataset provides labels: fraud and not-fraud. There are 433 features in the dataset. Most of the features are masked and their actual meanings are not provided due to privacy and security concerns. Therefore, we used 54 features for which the description of the features are known as the explanation could be more understandable. We now provide a brief explanation of the features that we kept. `TransactionAMT' gives the transaction amount in USD. `ProductCD' is the product category for each transaction. The device type such as mobile or desktop and information about the device such as operating system are given in `DeviceType' and `DeviceInfo' respectively. The features `card1'--`card6' are card related features such as debit or credit card type, card category (e.g., Mastercard, Visa, or American Express), issue bank, country, etc.  The purchaser's and receiver's email domains are given in `R\_emaildomain' and `P\_emaildomain' features respectively. `M1'--`M9' are match features, whether the names on the card and address, etc. match or not. `C1'--`C14' are count features such as how many addresses are found to be associated with the payment card. `D1'--`D15' are time delta features, such as days between previous transactions, etc. The address features are given in `addr1' and `addr2' and are, respectively, billing region (zip code) and billing country. `dist1' and `dist2' are distance features and are distance between billing address and mailing address. The following features are categorical features: `ProductCD', `card1'--`card6', `addr1', `addr2', `P\_emaildomain', `R\_emaildomain', `M1'--`M9' and the remaining features are numerical features. The data samples were divided into 60\% for training (i.e., 377K samples), 20\% (95K samples) for testing and 20\% (95K samples) for validation.  

\subsection{Classification Accuracy}
In this experiment, we compare the classification accuracy of ExMo, BRL and machine learning models on both small and large datasets. We used the AUC-ROC (area under ROC curve) score for comparison as used in~\cite{ieee_fraud,scalable_brl,brl}.  

\begin{table}
    \centering
    \caption{Classification scores on the small dataset}
    \label{tab:class_aucscore_uci_data}
    \begin{tabular}{ c | c | c | c}
     Dataset & BRL & XGBoost & Ours \\ \hline
     Diabetes & 0.66 &  0.79 & 0.81  \\ 
     Default Credit Card & 0.70 & 0.77 & 0.76 \\
     Income over 50K & 0.80 & 0.89 & 0.88 \\ \hline
    \end{tabular} 
\end{table}
\subsubsection{Small Dataset:}
Table~\ref{tab:class_aucscore_uci_data} shows the classification accuracy comparison on the small dataset for BRL, XGBoost and ExMo. Both ExMo and BRL were set to learn 10, 20 and 50 decision rules, respectively, for diabetes, default credit card payment and income over 50K datasets. The training was also executed for $K=30000$ iterations. Both ExMo and BRL were set to learn a maximum of 5 features in decision rules as well. The experimental result shows that ExMo achieves higher classification accuracy in all three datasets, thanks to TF-IDF based decision rules.  

\subsubsection{Large dataset:}
To evaluate the classification accuracy, we also build a deep learning model to detect fraud. The details of the deep learning model, algorithm settings and classification scores are provided below. 

\subsubsection{Deep Learning Model for Fraud Detection:}
We used a three-layered neural network to build the model for fraud detection. Embeddings were used for the categorical features and the output of the embedding layer is provided as input to the first layer. The first and second layers have 16 and 8 neurons, respectively, and the final layer is a sigmoid. The model outputs the probability of being a fraud for the given input sample.

\subsubsection{Algorithm Settings:}\label{exp:algo_settings}
We trained ExMo for three different $n$-gram settings of $(3, 5)$, $(5, 7)$, and $(7, 9)$ and these versions are referred, respectively, by `Ours-5', `Ours-7', and `Ours-9'. We used $P=200$ and $k=10$ in our experiments. With these settings, we extracted 3049, 3110, 3090 decision rules for the three different settings. Our algorithm is also configured to learn a model with a list of 125 decision rules. For the BRL algorithm, we used the maximum cardinality of 7 and the minimum support values of 75 and 65, respectively, for the negative and positive classes. With these settings, we extracted 3,600 decision rules. We also set the BRL algorithm to learn a model with a list of 125 decision rules. We used these settings for BRL because we got the best classification accuracy. 

\begin{table}
    \centering
    \caption{Classification scores on the IEEE-CIS Fraud dataset}
    \label{tab:class_aucscore_ieee_data}
    \begin{tabular}{ c | c }
     Method & AUC-ROC \\ \hline
     XGBoost & 0.93 \\ 
     Neural & 0.92  \\
     BRL & 0.69  \\ 
     Ours-5 & 0.82 \\ 
     Ours-7 & 0.87 \\  
     Ours-9 & 0.89  \\ \hline    
    \end{tabular} 
\end{table}
\subsubsection{Classification scores:}
Table~\ref{tab:class_aucscore_ieee_data} shows the classification accuracy comparison among different methods on test data.  We also provide results for three versions of ExMo: `Ours-5', `Ours-7', and `Ours-9'. We also implemented the XGBoost model that had won the competition in Kaggle \cite{ieee_fraud} and the result is given in the table for the comparison. The experimental results show that ExMo achieves a score that is very close to the deep neural network method and provides a significant improvement (20\%) over the BRL algorithm. The score saturates around `Ours-7', hence this model could be used in practice, which achieves a better trade-off of achieving higher accuracy and at the same time providing a compact explanation with fewer features. 

\subsection{Model Comparison}
In this section, we give the comparison between the models learnt by our algorithm and the BRL algorithm for the diabetes dataset. For diabetes, the plasma or blood glucose level is the main factor that indicates whether someone has diabetes or not. A High blood glucose level is a sign of diabetes. A blood glucose level below 95 is considered normal, 95-152 is considered pre-diabetic, and above 153 is considered diabetic~\cite{diabetes_ranges}. Therefore, the plasma glucose level is the primary feature whether some have diabetes or not. The higher value of body mass also increases the risk of being diabetic and a higher value not necessarily means diabetes. The risk of diabetes also increases with age. With these characteristics in mind, we will evaluate the model learnt by ExMo and BRL. 
\begin{table}[h]
    \centering
    \caption{Model comparison}
    \label{tab:model_text_diabetes}
    \begin{tabular}{ l }
    \hline
    \multicolumn{1}{c}{ExMo} \\ \hline
    \shortstack[l]{IF plasma\_glucose:127.5\_to\_166.5 AND body\_mass\_index:29.65\_to\_inf THEN \\ probability of diabetes: 76.0\% \\
    ELSE IF plasma\_glucose:127.5\_to\_166.5 THEN probability of diabetes: 41.5\% \\
    ELSE IF age:-inf\_to\_28.5 AND body\_mass\_index:-inf\_to\_29.65 THEN \\ probability of diabetes: 2.7\% \\
    ELSE IF plasma\_glucose:-inf\_to\_99.5 THEN probability of diabetes: 22.5\% \\
    ELSE IF plasma\_glucose:99.5\_to\_127.5 AND age:-inf\_to\_28.5 AND \\ times\_pregnant:-inf\_to\_6.5 THEN probability of diabetes: 30.1\% \\
    ELSE IF plasma\_glucose:99.5\_to\_127.5 THEN probability of diabetes: 60.9\% \\
    ELSE probability of diabetes: 54.1\%} \\ \hline
    \multicolumn{1}{c}{BRL} \\ \hline
    \shortstack[l]{IF age:28.5\_to\_inf AND body\_mass\_index:29.65\_to\_inf THEN \\ probability of diabetes: 74.9\% \\
    ELSE IF age:28.5\_to\_inf THEN probability of diabetes: 44.2\% \\
    ELSE IF serum\_insulin:-inf\_to\_16.0 AND body\_mass\_index:29.65\_to\_inf THEN \\ probability of diabetes: 59.6\% \\
    ELSE IF age:28.5\_to\_inf THEN probability of diabetes: 50.0\% \\
    ELSE IF body\_mass\_index:-inf\_to\_29.65 THEN probability of diabetes: 8.1\% \\
    ELSE IF plasma\_glucose:99.5\_to\_127.5 THEN probability of diabetes: 16.7\% \\
    ELSE IF body\_mass\_index:-inf\_to\_29.65 THEN probability of diabetes: 50.0\% \\
    ELSE IF age:-inf\_to\_28.5 THEN probability of diabetes: 50.6\% \\
    ELSE probability of diabetes: 50.0\%} \\ \hline
\end{tabular} 
\end{table}

Table~\ref{tab:model_text_diabetes} shows the model learnt by ExMo and BRL for the diabetes dataset. The decision rules in the ExMo model uses plasma glucose levels extensively, which is almost in every decision rule in the model compared to the decision rules in the model obtained by BRL. The first decision rule in ExMo uses plasma\_glucose in the range 127 to 166 and body\_mass\_index above 29 features to give a prediction of 0.76. On the other hand, the first rule in BRL uses age above 28 and body\_mass\_index above 29 features to give a prediction of 0.75, completely ignoring the plasma glucose level to give such a high prediction score. Subsequently, the decision rules in the model obtained by ExMo consisted of plasma glucose and other features such as age, body mass index, and the number of times the pregnancy had occurred. The model obtained by BRL consists of plasma glucose only on the 6th decision rule. Most of the decisions are based on age and body mass index. Theoretically, it is more accurate to look at the plasma glucose level as it is a primary feature and ExMo has been rightfully using this feature for prediction. Therefore, qualitatively, ExMo is able to learn a better model than BRL, thanks to TF-IDF based decision rules.    

\subsection{Textual Explanation}\label{sec:textual_explanation}
\begin{table}[]
    \centering
    \caption{Template for the textual explanation depending on rule types 1-4}
    \label{tab:explanation_template}
    \begin{tabular}{l|l}  \hline
    \multicolumn{2}{c}{Template for Explanation} \\ \hline
    \multicolumn{2}{l}{Prob. of \{application name\} is \{probability\} because} \\ \hline
    1. abs. value & \shortstack[l]{\{feature name\} is \{value\}} \\ \hline 
    2. -inf\_to\_value &  \{feature name\}(\{feature value\}) is below \{value\} \\ \hline
    3. value\_l\_to\_value\_r &  \shortstack[l]{\{feature name\}(\{feature value\}) is between \{value\_l\} \\ to \{value\_r\}} \\ \hline
    4. value\_to\_inf &  \{feature name\}(\{feature value\}) is above \{value\} \\ \hline
    \end{tabular}
\end{table}
In this section, we provide more details on how the textual explanation can also be generated, so that non-expert users can easily understand the explanation. Table~\ref{tab:explanation_template} shows the template that is used to format the explanation from the decision rule that was triggered. The 'application name' is the name of the application (e.g. diabetes) and the 'probability' is the output of the ExMo model. Depending on the feature types in the triggered decision rule, the templates from 1 to 4 given in Table~\ref{tab:explanation_template} are used to format the explanation. The first feature type is a feature with some absolute value, the second feature type considers a range of feature values from -infinity to some feature value, the third feature type considers a range of feature values from $value\_l$ to $value\_r$ and finally the fourth feature type considers a range of feature values from some feature value to infinity. The feature name and feature value are obtained from the input data sample.
\begin{table}[]
    \centering
    \caption{Data samples for a textual explanation}
    \label{tab:explanation_data_sample}
    \begin{tabular}{c|c|c|c|c|c}
    plasma{\_}glucose & serum{\_}insulin & body{\_}mass{\_}index & times{\_}pregnant & age & label \\ \hline
    152.0 & 171 & 34.2 & 9 & 33 & 1 \\ \hline
    108.0 & 0.0 & 30.8 & 2 & 21 & 0 \\ \hline
    \end{tabular}
\end{table}

Table~\ref{tab:text_explanation_pos_diabetes} shows the explanations generated by ExMo and BRL for the positive and negative data samples given in Table~\ref{tab:explanation_data_sample}. The template given in Table~\ref{tab:explanation_template} is used to convert the triggered decision rule into the textual explanation. For the positive sample, the first decision rule in the model for both ExMo and BRL in Table~\ref{tab:model_text_diabetes} is triggered. The explanation generated by ExMo states that the probability of diabetes is $0.76$ because the plasma glucose is in the higher range of 127 to 166 and body mass index is in the higher range and above the threshold value of 29. Similarly, for BRL, the first decision rule is triggered and the textual that is obtained using the template is shown in the table. For the negative sample, the 5th and 3rd rules are triggered, respectively for ExMo and BRL. The textual explanation generated by ExMo states that the probability of diabetes is $0.3$ because the plasma glucose is between 99.5 to 127.5, the patient is young in the age of 21 which is below 28 and the number of pregnancies that occurred so far is 2.0 and below some threshold of 6.5. It is quite clear that the textual explanation is easily understandable and readable by non-experts.
\begin{table}[]
    \centering
    \caption{Textual explanation of positive and negative samples in the diabetes dataset}
    \label{tab:text_explanation_pos_diabetes}
    \begin{tabular}{l|l}
     Method    &  \multicolumn{1}{c}{Textual Explanation} \\ \hline
     \multicolumn{2}{c}{Positive sample (label 1)}   \\ \hline
     BRL      & \shortstack[l]{Prob. of diabetes is 0.75 because age (33.0) is above 28.5 and \\ body{\_}mass{\_}index (34.2) is above 29.65} \\ \hline
     Ours     & \shortstack[l]{Prob. of diabetes is 0.76 because plasma{\_}glucose (152.0) is between \\ 127.5 to 166.5 and body{\_}mass{\_}index (34.2) is above 29.65}    \\ \hline
     \multicolumn{2}{c}{Negative sample (label 0)}   \\ \hline
     BRL      & \shortstack[l]{Prob. of diabetes is 0.6 because serum{\_}insulin (0.0) is below 16.0 and \\ body{\_}mass{\_}index (30.8) is above 29.65} \\ \hline
     Ours     & \shortstack[l]{Prob. of diabetes is 0.3 because plasma{\_}glucose (108.0) is \\ between 99.5 to 127.5 and age (21.0) is below 28.5 and \\ times{\_}pregnant (2.0) is below 6.5}   \\ \hline
    \end{tabular}
\end{table}

\subsection{Comparison with Model-Agnostic Explanation}
In this section, we compare the explanations generated by our method, BRL and model agnostic explanation methods such as SHAP, LIME, Anchor and IG. We used the large dataset of fraud detection for the experiment. For the model agnostic methods, we also need a deep learning model to explain the decision. Hence, we used a deep neural network for fraud detection as discussed earlier. 

\subsubsection{Algorithm Settings}\label{exp:algo_settings}
We used the implementation provided by the authors for SHAP~\cite{shap_code}, LIME~\cite{lime_code}, Anchor~\cite{anchor_code} and BRL~\cite{brl}. For Integrated Gradients, we used the implementation provided in~\cite{alibi}. For SHAP, we also needed to provide the background samples. We selected all fraud samples and the equal number of the non-fraud samples from the training data as the background samples. Since the dataset is unbalanced, providing the background samples in this way gives more stable explanations in our experiments. 

\subsubsection{Common Features in Explanations:}
\begin{figure*} 
    \centering
    \includegraphics[width=4.80in]{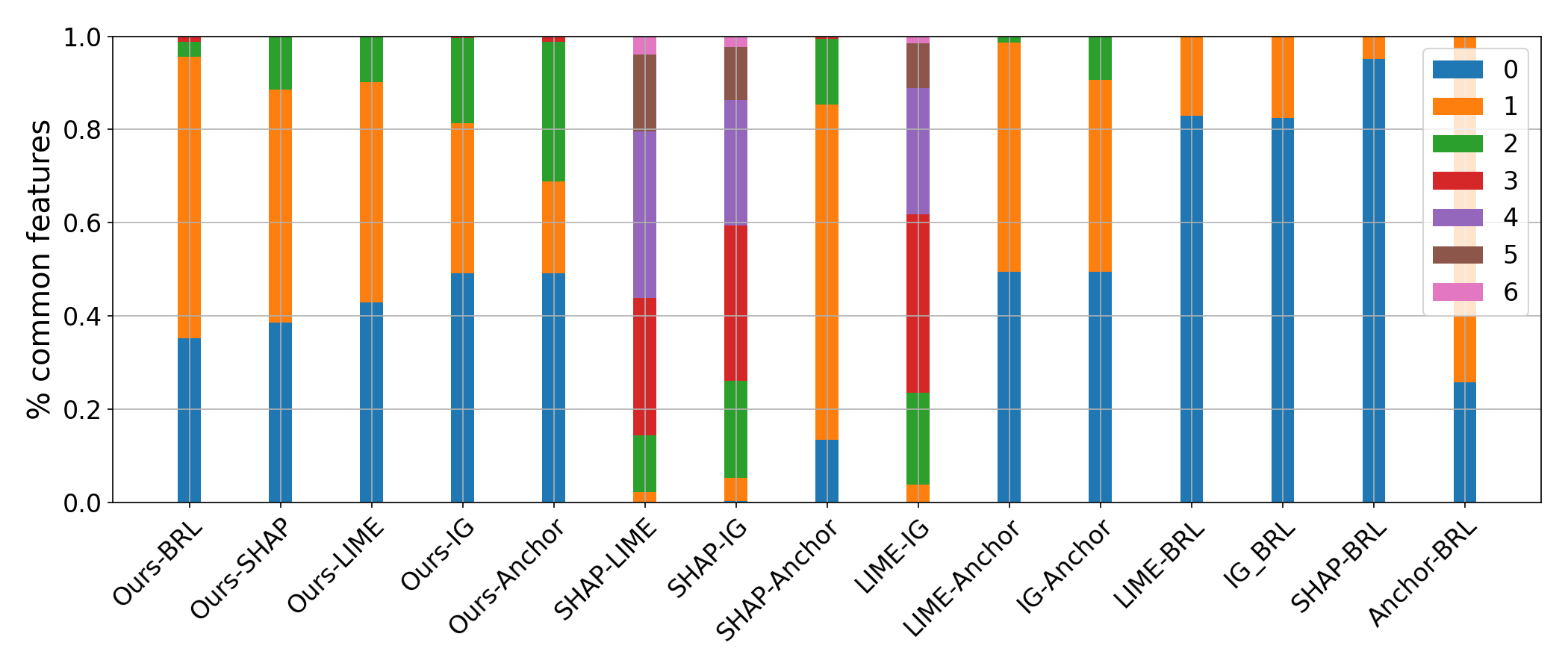}
    \caption{The number of features that are common in explanation generated by the different algorithms.}
    \label{fig:common_features}
\end{figure*}

In this experiment, we report how many matching features are found in the explanation given by different algorithms. Ideally, we would like to have same features in the explanation for all the algorithms. In this experiment, we used 10 thousand samples from the test set. All 4 thousand fraud samples and the remaining randomly sampled 6 thousand non-fraud samples are used. We used the `Ours-7' version of our algorithm, which generates at most 7 features in explanation. The remaining algorithms (SHAP, LIME, IG, and Anchor) used in the comparison are also set to generate 7 features in the explanation. The number of matching features in the explanations are counted. Fig.~\ref{fig:common_features} shows a distribution of counts of features that are common in the explanations given by the different combinations of the algorithms. The result shows that ExMo finds significantly more matching features with other methods in explanation as compared to the BRL algorithm. For the BRL algorithm, around 85\% of the samples did not have any matching features with the explanation generated by LIME, IG, and SHAP. For ExMo, only around 40\% of the samples did not have any common features in the explanation.

\subsubsection{Instance Explanations:}
\begin{figure*}
\centering
    \subfloat[]{\includegraphics[width=2.4in]{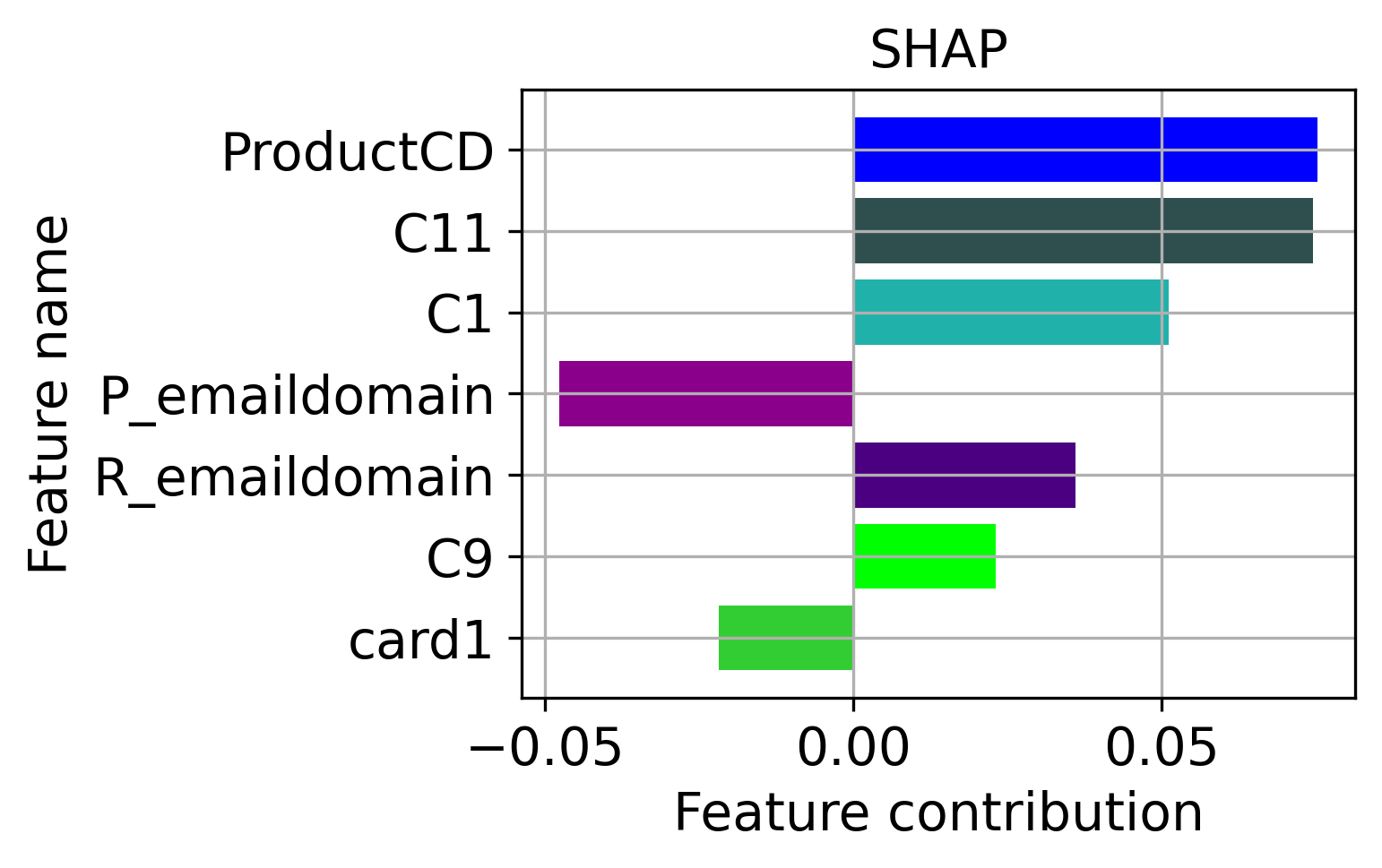}}
    \subfloat[]{\includegraphics[width=2.4in]{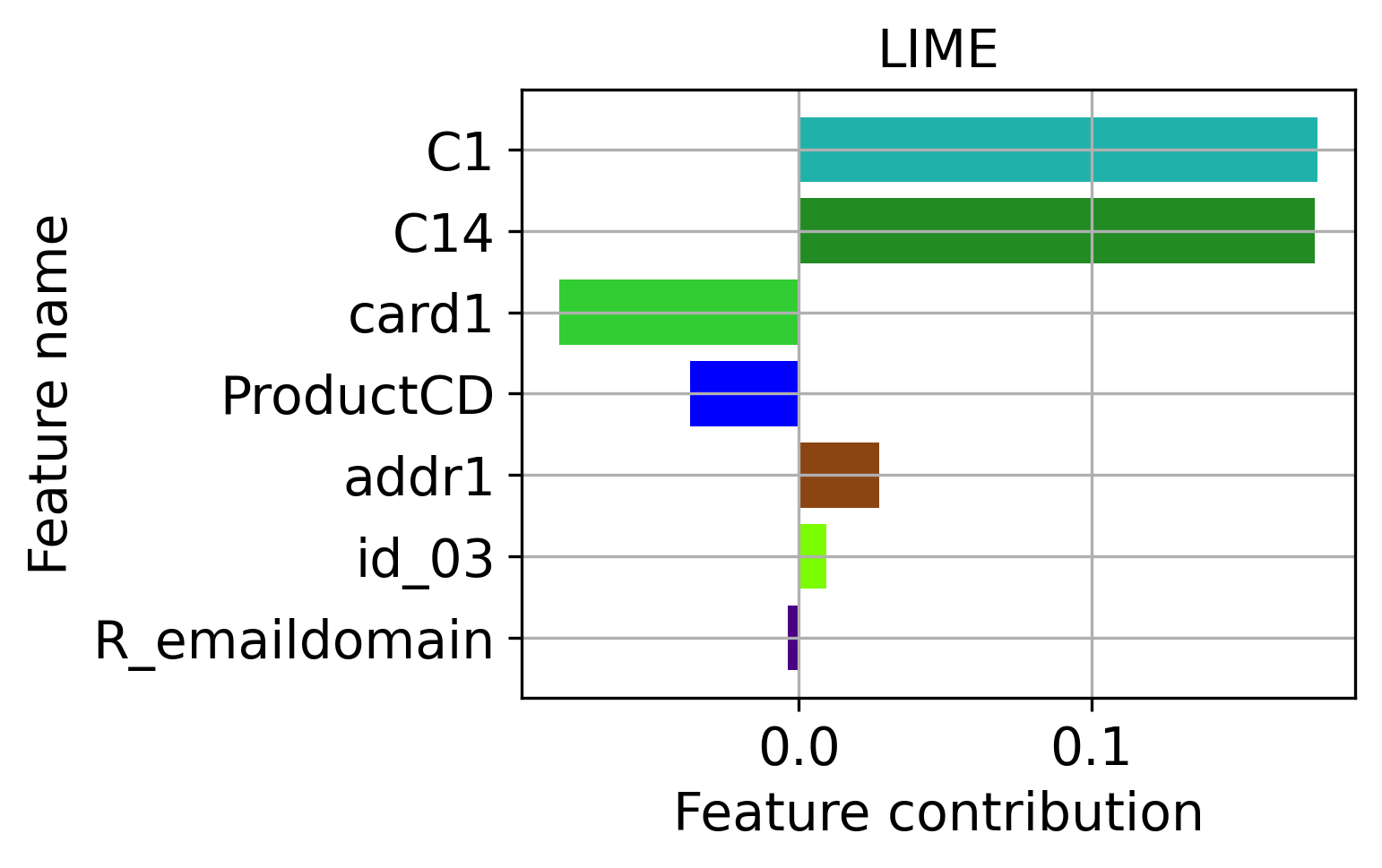}} \\
    \subfloat[]{\includegraphics[width=2.4in]{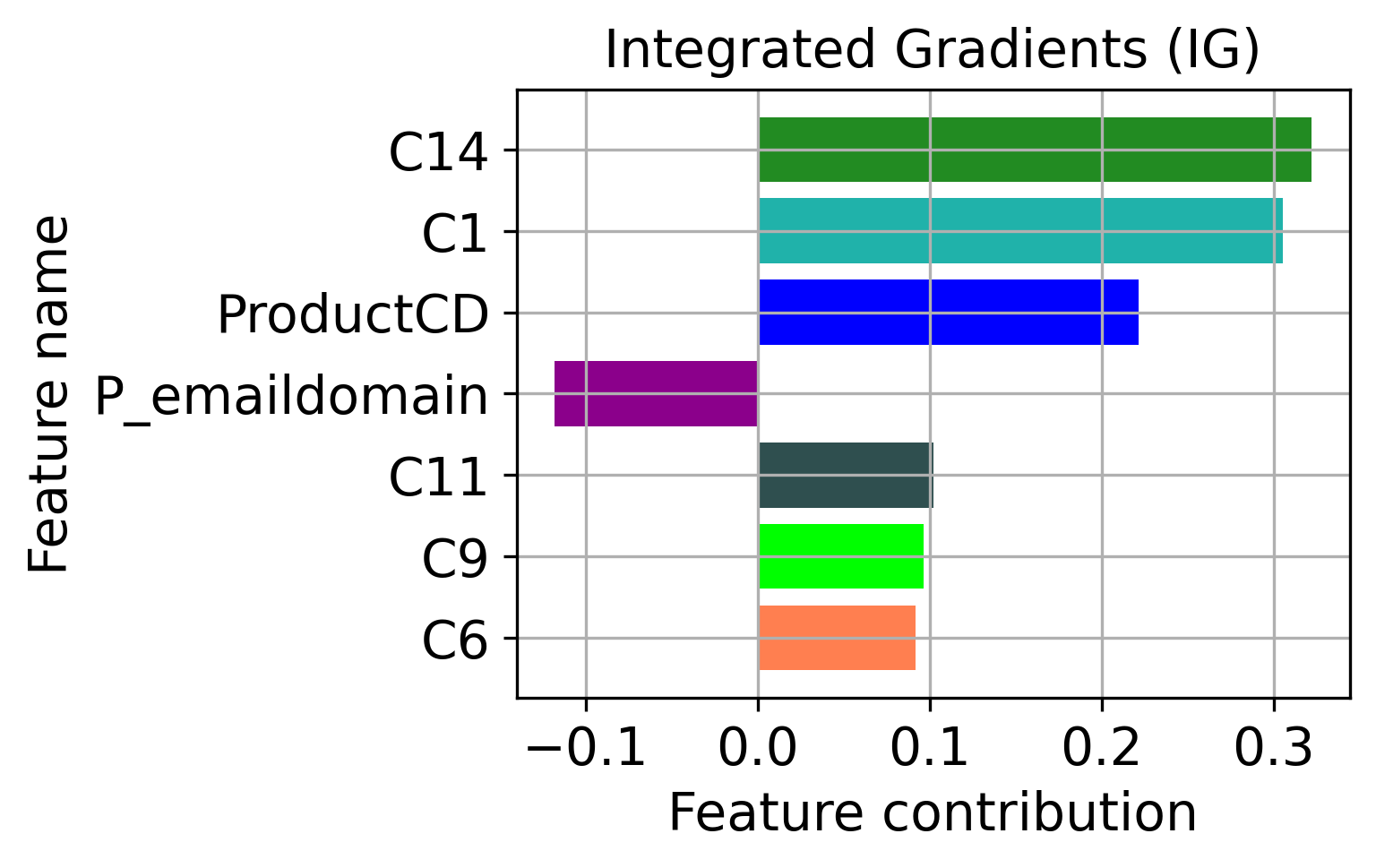}}  
    \subfloat[]{\includegraphics[width=2.4in]{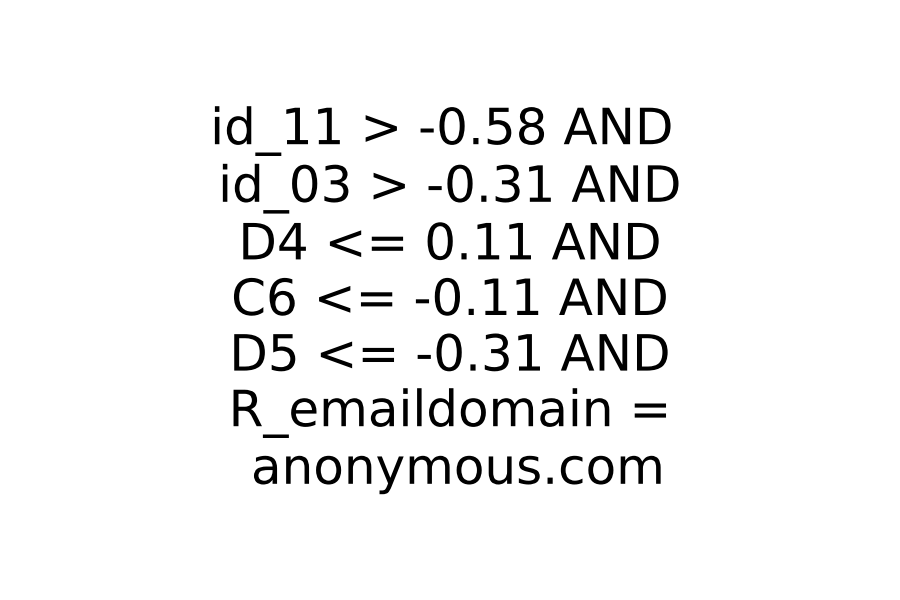}} \\
    \subfloat[]{\includegraphics[width=2.4in]{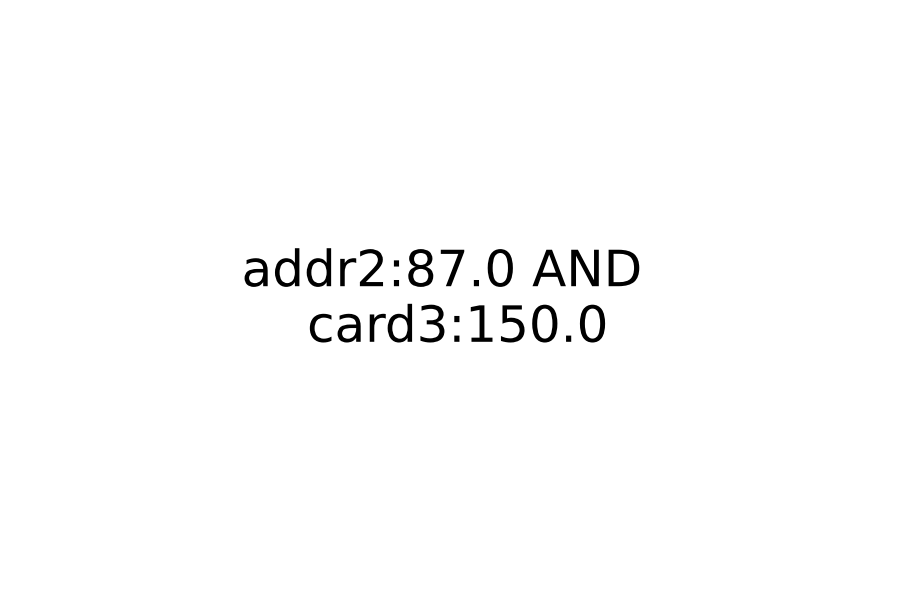}}
    \subfloat[]{\includegraphics[width=2.4in]{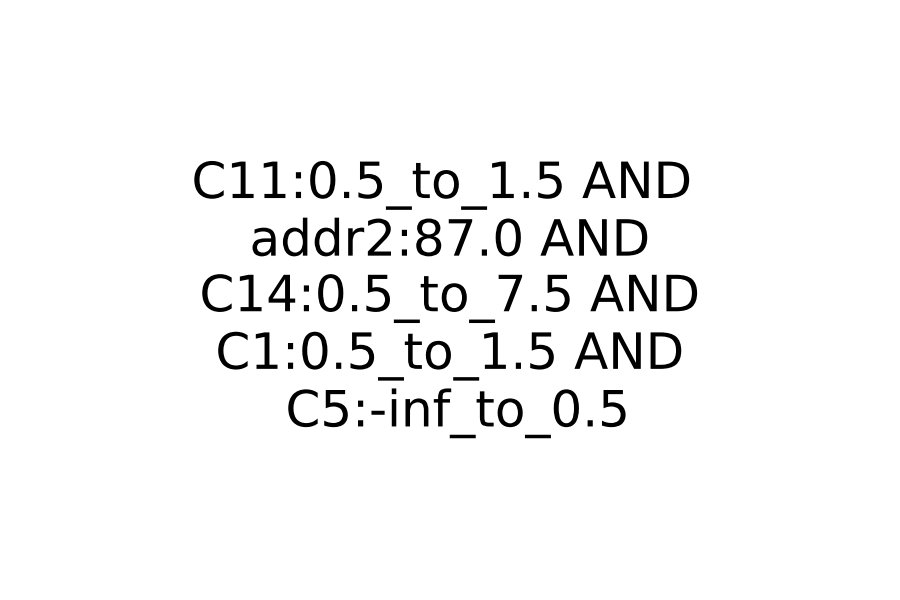}}
    \caption{Explanations computed for a non-fraud sample by: (a)~SHAP, (b)~LIME, (c)~Anchor, (d)~Integrated Gradients, (e)~BRL and (f)~Ours. The values of the features that appeared in the explanations are: ProductCD: S, C1: 1.0, C5: 0.0, C6: 0.0, C9: 0.0, C11: 1.0, C14: 1.0, P\_emaildomain: NULL, R\_emaildomain: anonymous.com, card1: 1675, card3: 150.0, aadr1: 330.0, addr2: 87.0, id\_03: 0.0, id\_11: 100.0, D4: NULL, D5: NULL.}
    \label{fig:instance_explanations_notfraud}
\end{figure*}

\begin{figure*}
\centering
    \subfloat[]{\includegraphics[width=2.4in]{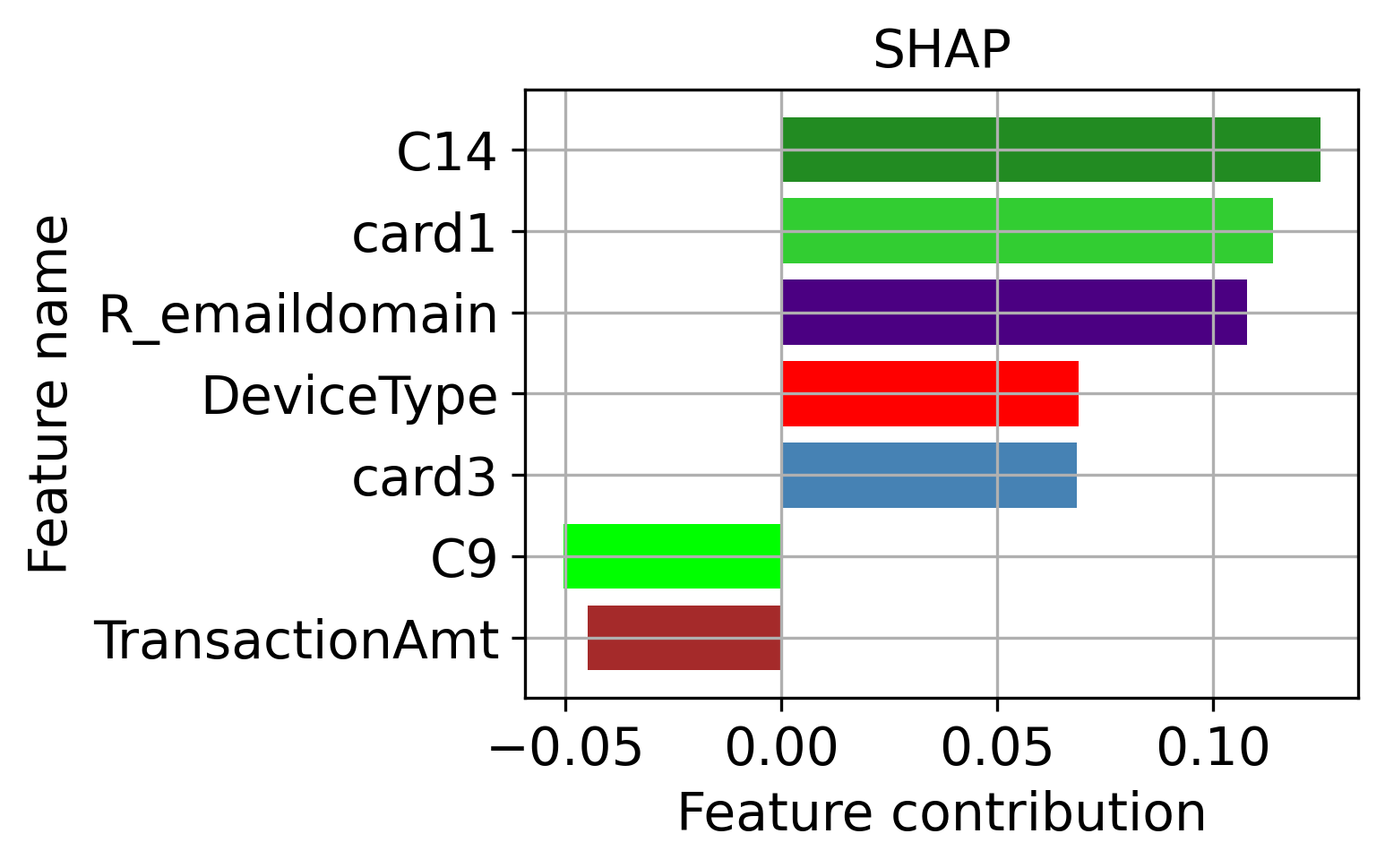}}
    \subfloat[]{\includegraphics[width=2.4in]{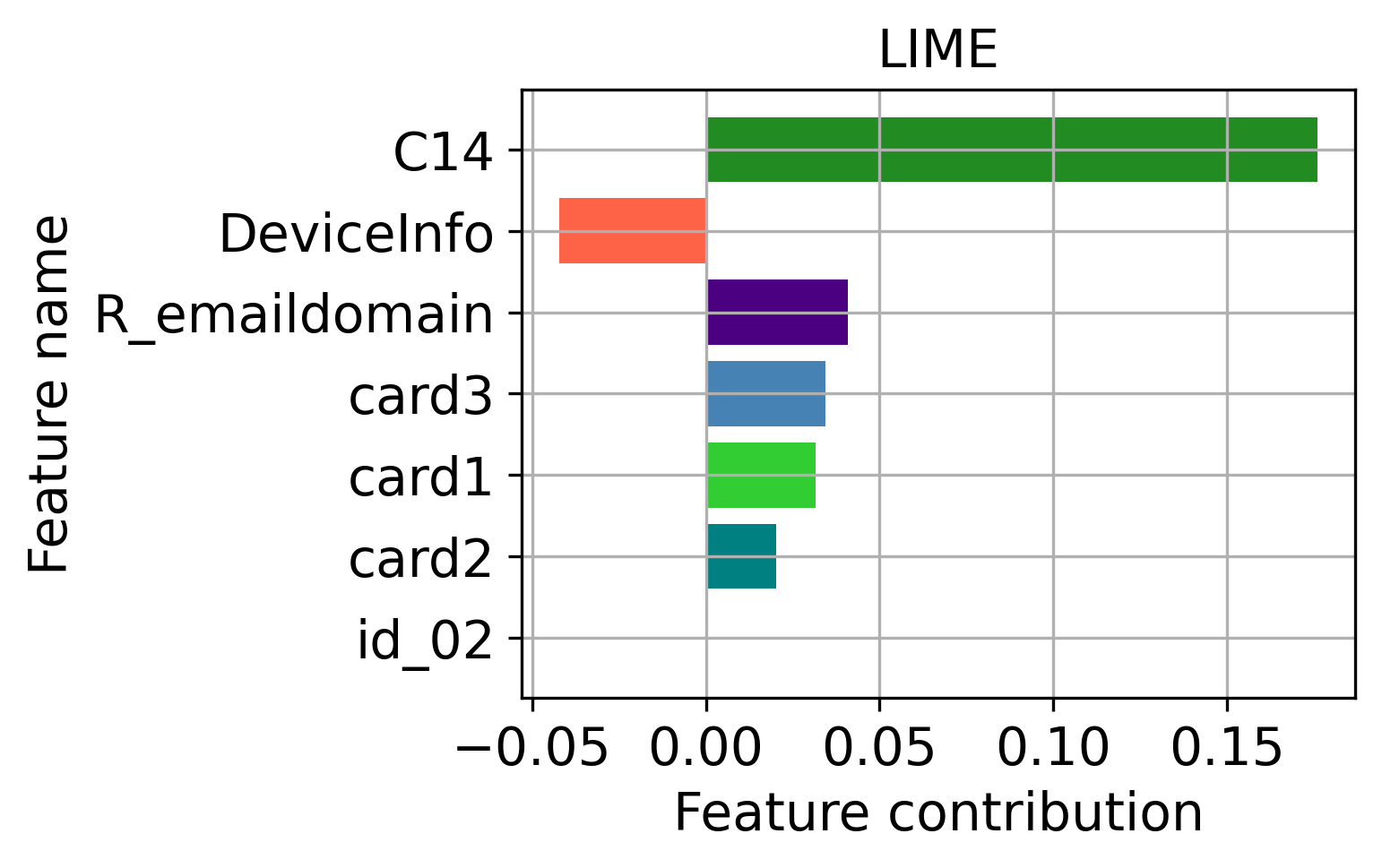}} \\
    \subfloat[]{\includegraphics[width=2.2in]{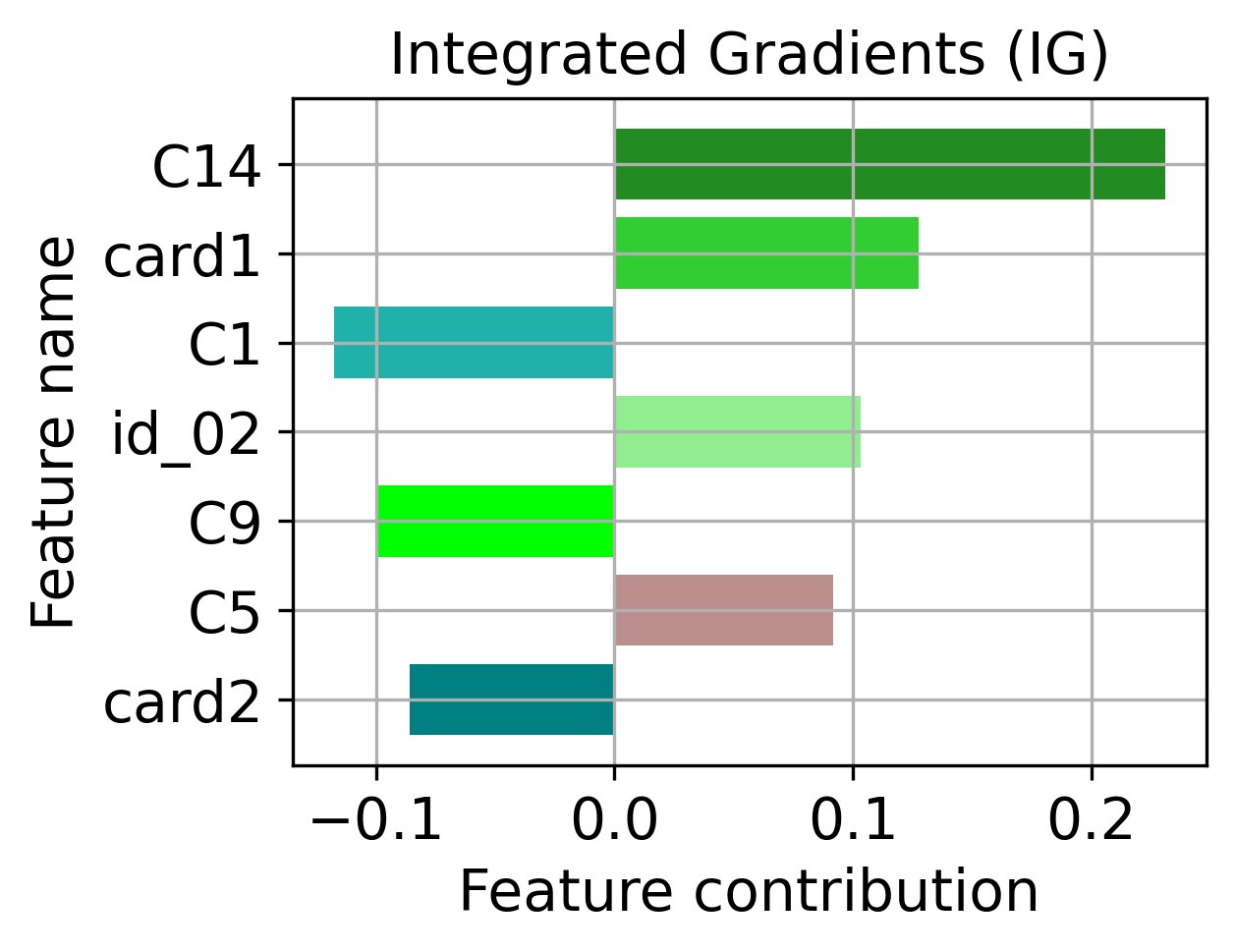}} 
    \subfloat[]{\includegraphics[width=2.4in]{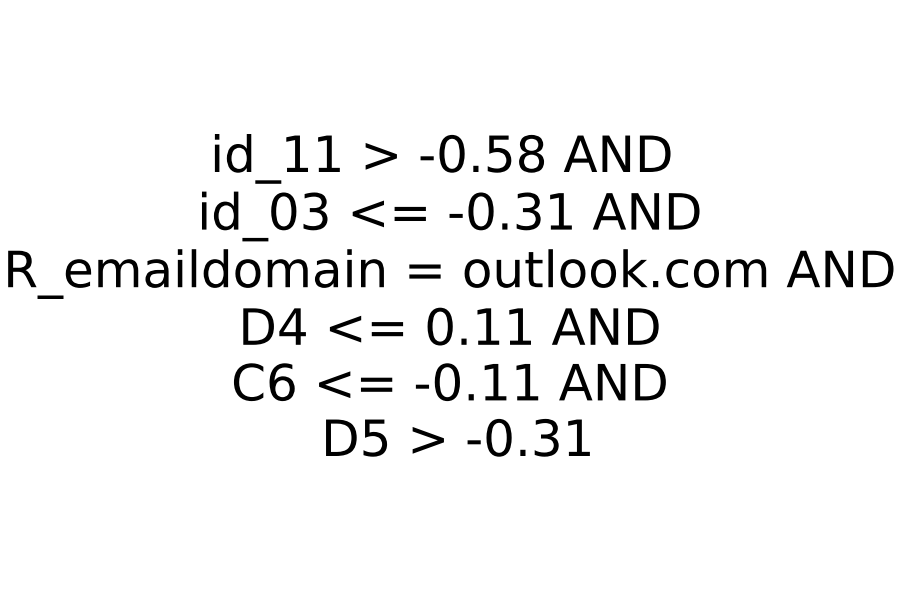}} \\
    \subfloat[]{\includegraphics[width=2.4in]{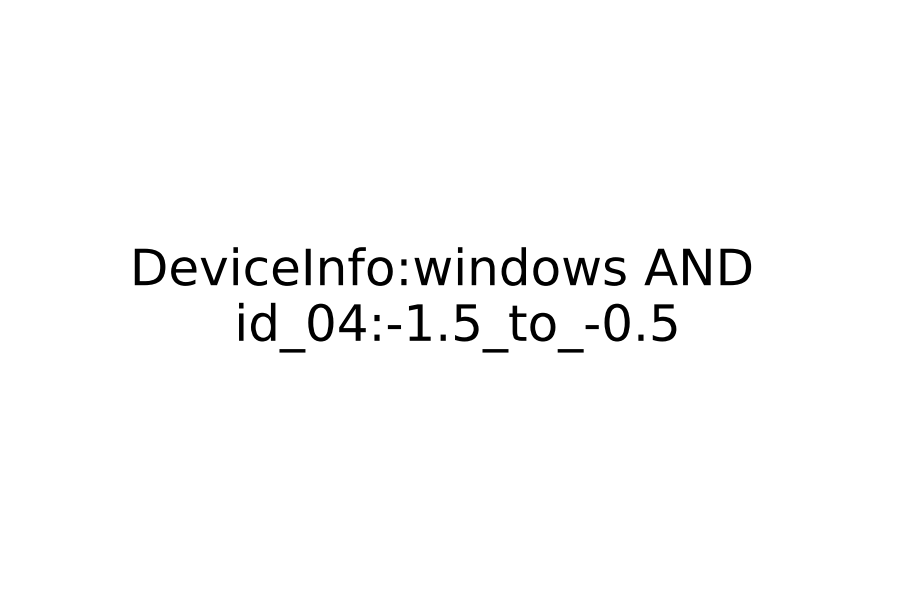}}
    \subfloat[]{\includegraphics[width=2.4in]{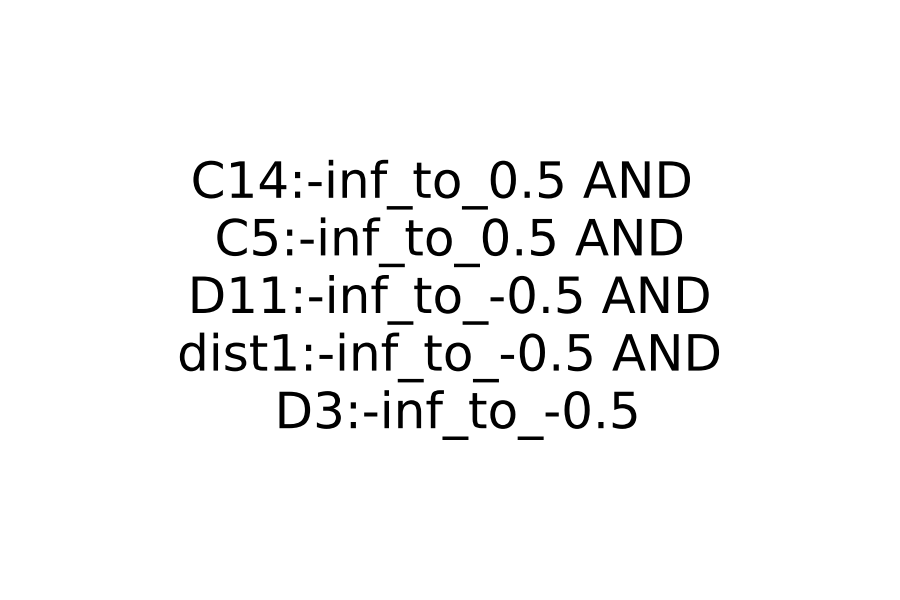}}
    \caption{Explanations computed for a fraud sample by: (a)~SHAP, (b)~LIME, (c)~Anchor, (d)~Integrated Gradients, (e)~BRL and (f)~Ours. The values of the features that appeared in the explanations are: ProductCD: C, C1: 3.0, C5: 0.0, C6: 1.0, C9:0.0, C14: 0.0, TransactionAmt: 36.581, R\_emaildomain: outlook.com, card1: 9633, card2: 130.0, card3: 185.0, DeviceInfo: NULL, DeviceType: mobile, dist1: NULL, id\_02: 522232.0, id\_03: NULL, id\_04: NULL, D3: NULL, D4: 0.0, D5: 0.0, D11: NULL.}
    \label{fig:instance_explanations_fraud}
\end{figure*}

In this section, we provide the results of the explanation generated by different algorithms. Figures~\ref{fig:instance_explanations_notfraud} and~\ref{fig:instance_explanations_fraud} show explanations generated by SHAP, LIME, IG, Anchor, BRL, and our approach for non-fraudulent and fraudulent transactions, respectively. The model predictions for the non-fraudulent transaction used in Fig.~\ref{fig:instance_explanations_notfraud} are 0.005, 0.512, and 0.345, respectively, for the deep learning model, BRL and ExMo. In this example, two and three common features in the explanation are found by ExMo and SHAP and IG, respectively. The features `C11' and `C1' are common in the explanation given by SHAP and ExMo. The three features `C14', `C11', and `C1' are common in the explanation given by IG and our approach. For the fraudulent transaction given in Fig.~\ref{fig:instance_explanations_fraud}, the predictions are 0.70, 0.89, and 0.95 for the deep learning model, BRL and ExMo. There are one and two common features in the explanation for SHAP and IG with ExMo, respectively.  

\subsubsection{Discussion:}
SHAP, LIME and IG provide explanations in terms of feature importance, providing positive and negative contributions. For example, the SHAP explanation shown in~\ref{fig:instance_explanations_notfraud}(a) gives the importance of each feature `ProductCD', `C11', `C1', etc. that contribute positively to the prediction. The feature explanation only provides information in terms of importance. In our approach, BRL and Anchor provide explanations in the form of a decision rule, which is different from the feature importance. The explanation is given in terms of comparison, i.e., if the feature value is greater than or equal to some value. Therefore, the explanation of the decision rule is in the absolute sense (i.e., pointing out the feature value or feature value range), while the explanations in the form of feature importance are in the relative sense in terms of percentage of contribution. Therefore, the textual explanation as we discussed in Sec.~\ref{sec:textual_explanation} will not be possible for model agnostics explanation techniques as only feature importance's are known. One thing to note is that the features in the decision rule do not give importance to any features and also they are not presented in the sequence of their importance. 

\subsection{Execution Time}\label{sec:execution_time}
Table~\ref{tab:execution_time} shows execution time in seconds to compute the explanation for the IEEE-CIS fraud dataset. The experiments were carried out on a Linux server with Intel\textsuperscript{\textregistered} Xeon\textsuperscript{\textregistered} Silver 4114 CPU running at 2.20GHz. ExMo and BRL are computationally very efficient compared to the other methods. Both methods only need to discretize the numerical inputs and then perform comparisons with the decision rules. Both prediction and explanation are computed very efficiently. Anchor has the largest computational overhead. The computational efficiency of ExMo can be particularly useful in real-time use cases that need to provide a prediction and explanation within specific time limits.
\begin{table}
    \centering
    \caption{Execution time to compute prediction and explanation in seconds}
    \label{tab:execution_time}
    \begin{tabular}{l|r}
    \multicolumn{1}{c} {Method} & {Execution Time} \\ \hline
     Ours-7 & 0.41 \\
     BRL & 0.42 \\
     IG & 3.22 \\
     LIME & 3.25 \\
     SHAP & 5.21  \\
     Anchor & 67.25 \\ \hline
    \end{tabular} 
\end{table}

\section{Conclusion}
In this paper, we proposed a new method to compute decision rules and build a more accurate interpretable machine learning model, named ExMo. ExMo uses the TF-IDF algorithm to extract decision rules, hence we are able to extract higher-quality decision rules compared to the frequent pattern mining approach used in the original BRL algorithm. The efficacy of ExMo is also validated on several datasets with different sample and feature sizes. The validation of ExMo on fraud detection, which is a very complex and challenging problem for machine learning systems, shows that ExMo can achieve accuracy very close to the black-box deep learning method. We also demonstrated with the experiments that with the use of these decision rules on fraud detection applications that ExMo can achieve $\approx$20\% higher accuracy. We also demonstrated that the textual explanation can be provided by ExMo, which could be used to communicate decisions to non-expert users. We conclude that our approach will be useful for high-stake applications since it not only achieves better accuracy but also provides explanations in a more human-friendly way. An additional advantage is that the model can also be inspected by regulators for bias and fairness, in a way that is user friendly and easy to understand.

%
% ---- Bibliography ----
%
% BibTeX users should specify bibliography style 'splncs04'.
% References will then be sorted and formatted in the correct style.
%
\bibliographystyle{splncs04}
\bibliography{my_bib}

\end{document}